%% file: main.tex
\documentclass[10pt,twocolumn,letterpaper]{article}

\usepackage{cvpr}              
\input{preamble}

\definecolor{cvprblue}{rgb}{0.21,0.49,0.74}
\usepackage[pagebackref,breaklinks,colorlinks,citecolor=cvprblue]{hyperref}

\def\paperID{5941} 
\def\confName{CVPR}
\def\confYear{2025}

\title{MROVSeg: Breaking the Resolution Curse of Vision-Language Models in Open-Vocabulary Image Segmentation}

\author{Yuanbing Zhu$^{1,2}$, Bingke Zhu$^{1}$, Yingying Chen$^{1}$, Yunfang Niu$^{1}$, Ming Tang$^{1,2}$, Jinqiao Wang$^{1,2}$\\ 
\normalsize $^1$Foundation Model Research Center, Institute of Automation, Chinese Academy of Sciences \\
\normalsize $^2$School of Artificial Intelligence, University of Chinese Academy of Sciences \\
{\tt\small \{bingke.zhu, yingying.chen, jqwang, tangm\}@nlpr.ia.ac.cn} \quad
{\tt\small zhuyuanbing2021@ia.ac.cn} 
}

\begin{document}
\maketitle
\input{sec/sec_0_abstract}
\input{sec/sec_1_introduction}
\input{sec/sec_2_related_works}
\input{sec/sec_3_method}
\input{sec/sec_4_experiments}

\input{sec/sec_5_conclusion}
{
    \small
    \bibliographystyle{ieeenat_fullname}
    \bibliography{main}
}
\end{document}

%% file: preamble.tex
\usepackage[dvipsnames]{xcolor}

\usepackage[dvipsnames]{xcolor}
\usepackage{multirow}
\usepackage{bbding}
\usepackage{subcaption}
\usepackage{colortbl}
\usepackage{algorithm}
\usepackage{booktabs}
\usepackage{algorithmic}

\usepackage{newfloat}
\usepackage{listings}
\DeclareCaptionStyle{ruled}{labelfont=normalfont,labelsep=colon,strut=off} 
\floatstyle{ruled}
\newfloat{listing}{tb}{lst}{}
\floatname{listing}{Listing}

%% file: sec/sec_0_abstract.tex
\begin{abstract}
Pretrained vision-language models (VLMs), \eg CLIP, are increasingly used to bridge the gap between open- and close-vocabulary recognition in open-vocabulary image segmentation. As VLMs are generally pretrained with low-resolution images (e.g. $224\times224$), most previous methods operate only on downscaled images. We question this design as low resolution features often fail to preserve fine details. A typical solution is to employ additional image backbones for high-resolution inputs, but it also introduce significant computation overhead. Therefore, we propose MROVSeg, a multi-resolution training framework for open-vocabulary image segmentation with a single pretrained CLIP backbone, that uses sliding windows to slice the high-resolution input into uniform patches, each matching the input size of the well-trained image encoder. Its key components include a Multi-Res Adapter, which restores the spatial geometry and grasps local-global correspondences across patches by interacting with multi-resolution features. To achieve accurate segmentation, we introduce Multi-grained Masked Attention scheme to aggregate multi-grained semantics from multi-resolution CLIP features to object queries. Through comprehensive experiments, we demonstrate the superiority of MROVSeg on well-established open-vocabulary image segmentation benchmarks, establishing new standards for open-vocabulary image segmentation. 
\end{abstract}

%% file: sec/sec_1_introduction.tex
\section{Introduction}
\label{sec:intro}
\begin{figure*}[t!]
  \centering
    \begin{subfigure}[b]{0.47\linewidth}
      \centering
      \includegraphics[width=1.0\linewidth]{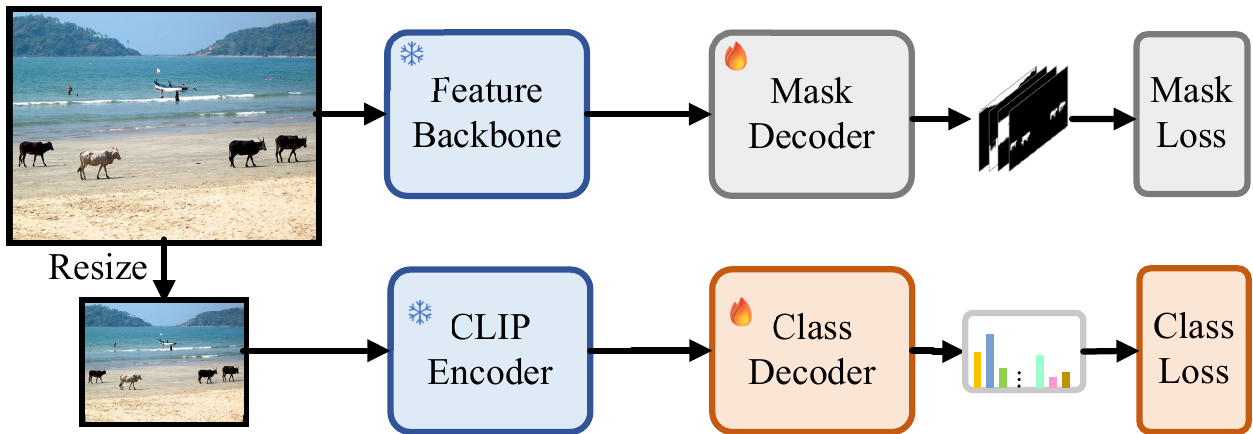}
      \caption{\textbf{Previous open-vocabulary image segmentation.}}
      \label{fig:intro:1_a}
     \end{subfigure}
    \hfill
    \begin{subfigure}[b]{0.47\linewidth}
      \centering
      \includegraphics[width=1.0\linewidth]{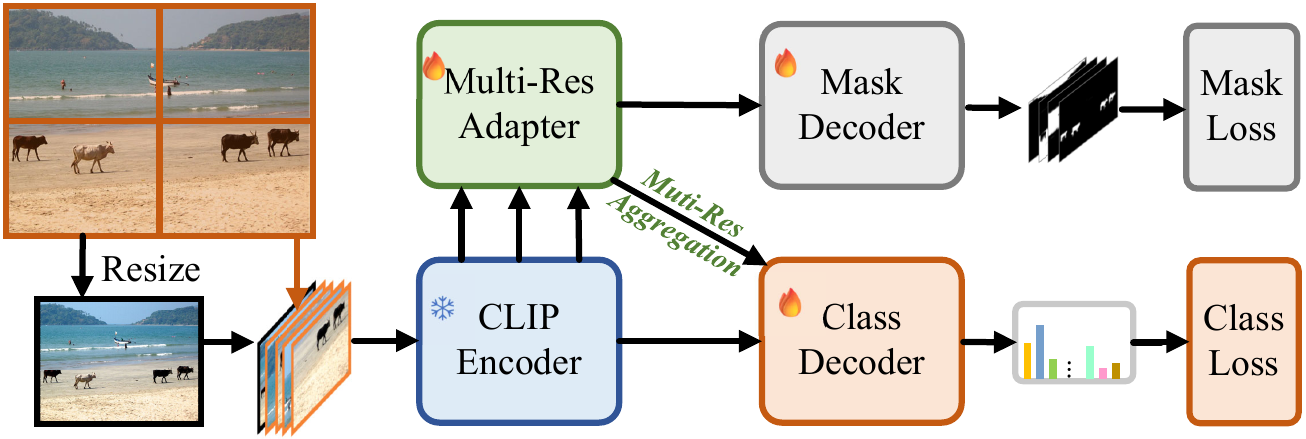}
      \caption{\textbf{MROVSeg open-vocabulary image segmentation.}}
      \label{fig:intro:1_b}
     \end{subfigure}
    \caption{\textbf{Comparison between other training frameworks and MROVSeg.} Previous methods (a) adopt additional image backbone to provide mask feature. The mask prediction is \textit{class-unaware}. Our method (b) provide multi-resolution CLIP feature for both mask decoding and mask classification, and the whole framework is \textit{class-aware}.}
  \label{fig:1}
\end{figure*}
Open-vocabulary image segmentation aims to segment semantic pixels belonging to arbitrary classes beyond pre-defined categories and dataset. In recent years, large-scale vision-language pretrained models (VLMs), such as CLIP~\cite{clip} and ALIGN~\cite{align}, have demonstrated remarkable generalization capabilities for recognizing open-vocabulary categories. This motivates the research community to investigate the potential of VLMs in open-vocabulary image segmentation. To address the discrepancy between the per-pixel semantic requirements and the image-level labels provided by VLMs, initial studies~\cite{catseg,denseclip,zegclip} modified CLIP model by removing its final pooling layer to obtain dense category embeddings for per-pixel classification. However, these approaches typically necessitate fine-tuning VLMs on a base segmentation dataset with limited images and categories, which is demonstrated~\cite{denseclip} to impair the transferability of VLM features, leading to unsatisfactory zero-shot performance on downstream tasks.

Recent approaches~\cite{simbase,maskclip,san,ebseg,scan} reformulate open-vocabulary image segmentation as a region-level recognition problem. These methods typically adopt two branch meta architecture (As in \cref{fig:intro:1_a}): one branch extract image feature and generate mask proposals, and the other branch classifies the predicted proposals with pretrained VLM. Although these methods are promising, we note
their following limitation. Due to pretrained VLMs exhibit inferior size adaptability, most of open-vocabulary image segmentation methods (e.g. \cite{maskclip,san,ovseg,simbase,odise,opsnet,ebseg, scan}) so far need to downsample images to fit the pretrained resolution (e.g. $224\times224$) of VLM to perform region-level recognition (as in \cref{fig:intro:1_a}). However, low-resolution input usually lacks segmentation details. Although na\"ively applying sliding window inference~\cite{catseg, simbase} could partly compensate for the details, the spatial structure across windows is corrupted and the local-global modeling is also absent.

In light of the limitations and challenges faced by previous methods, we propose MROVSeg, a VLM-based \textbf{M}ulti-\textbf{R}esolution training framework for \textbf{O}pen-\textbf{V}ocabulary Image \textbf{Seg}mentation. As illustrated in ~\cref{fig:intro:1_b}, first, MROVSeg uses downsampled low-resolution images as VLM input to extract global low-resolution features. Second, MROVSeg split the high-resolution images into slices and feeds them to VLM to extract detailed high-resolution features. The key components of MROVSeg contain a Multi-Res Adapter, in which we employs depthwise convolution layers to restore the spatial geometry across slices. To effectively capture global long-range context, inspired by previous multi-scale training framework~\cite{scaleattn,hoyer2022hrda,msca}, we employ a image-dependent Scale-aware Attention~\cite{scaleattn} to dynamically adjust the trustworthiness of high-resolution and low-resolution VLM features based on their relevance. The resulting multi-resolution features are fused hierarchically then employed for precise mask proposals generation.

To achieve accurate mask class recognition, we propose a Multi-grained Masked Attention mechanism. The core hypothesis is that multi-resolution CLIP features of the save image input hold semantic consistency. Based on this, we reuse the CLIP \texttt{[CLS]} token, and manipulate its attention map on multi-resolution features in CLIP attention layers with \textit{resolution-aware} attention masks. We find this resolution-aware design can enforce the low- and high-resolution attention map focus on global contexts and spatial details respectively and thus effectively aggregate multi-grained semantics.
With extensive experiments on well-established open-vocabulary semantic segmentation and panoptic segmentation benchmarks, we are delighted to report that our method achieves new state-of-the-art performance, demonstrating the advancements of MROVSeg in the domain of open-vocabulary image segmentation. Our contributions can be summarized as follows:
\begin{itemize}
    \item We propose a novel end-to-end multi-resolution training framework to tackle the task of open-vocabulary image segmentation. It enables improved open-vocabulary segmentation by leveraging multi-resolution vision-language features.
    \item A multi-grained masked attention scheme is proposed to effectively aggregate regional and universal semantics on multi-resolution vision-language features.
    \item The efficacy of our method is confirmed by achieving state-of-the-art performance by evaluating our proposed approach on well-established open-vocabulary segmentation benchmarks.
\end{itemize}

%% file: sec/sec_2_related_works.tex
\section{Related work}
\label{sec:related_work}
\begin{figure*}[t]
  \centering
    \includegraphics[width=0.99\linewidth]{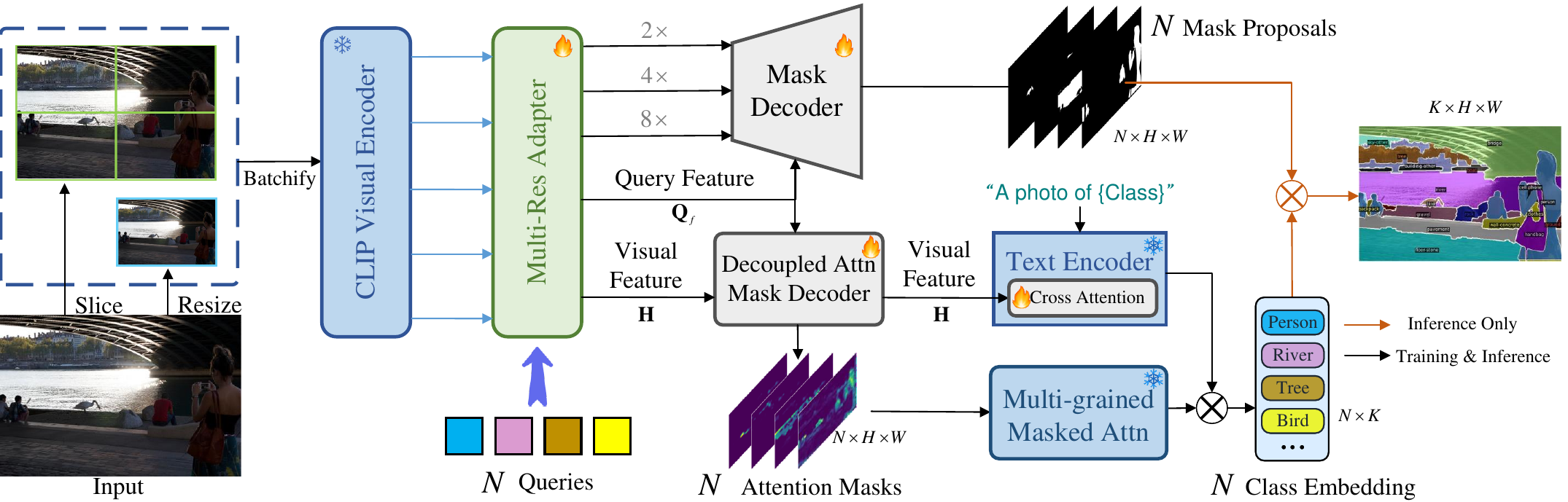}
    \caption{\textbf{The overall pipeline of MROVSeg.} For an high-resolution input image, its downsampled image and  are fed into CLIP visual encoder to extract multi-resolution CLIP features. The Multi-Res Adapter adapts these features for mask decoder and attention mask decoder. The generated attention masks are employed to aggregate semantics from the multi-resolution CLIP features.}
    \label{fig:overall_architecture}
  \hfill
  \vspace{-10pt}
\end{figure*}
\noindent
\textbf{Pretrained Vision-language Models}\quad Recently, large-scale pretrained, contrastive learning based methods~\cite{clip,align} demonstrate powerful open-vocabulary image classification capability by learning shared image-text feature representations. Pretrained CLIP~\cite{clip} has been generalized to many downstream computer vision tasks such as object detection~\cite{regionclip}, image caption~\cite{clipscore}, image generation~\cite{styleclip} and image segmentation~\cite{san, zs3net,odise,fcclip,simbase}. Our method invokes natural language perceptual ability of pretrained VLMs, aiming to explore their application boundary in open vocabulary semantic segmentation tasks. \\
\textbf{Multi-Resolution Training}\quad As the quadratic computational overhead along with the number of tokens increases, recent multimodal large language models~\cite{llava15,li2024monkey, xu2024llava} (MLLMs) employ sliding window technique to divide high-resolution images into patches, thereby achieving competitive performance while maintaining computational efficiency. Unlike prevalent MLLMs, MROVSeg adaptively restores the spatial geometry of multi-resolution features across patches, and effectively extract global contexts that benefit segmentation tasks.\\
\noindent
\textbf{Open Vocabulary Image Segmentation}\quad Pioneering works~\cite{zs3net,spnet} use learned language embedding to align the feature space of class texts and visual semantics. Recently, some works~\cite{simbase, maskclip, ovseg, scan} develop two-stage training approaches. More recently, end-to-end~\cite{san} frameworks emerge in the community, which unify mask generation and region classification into the same model. SAN~\cite{san} propose a lightweight side adapter to effectively adapt CLIP features. ODISE~\cite{odise} employs a stable diffusion UNet~\cite{stablediff} to generate mask proposals. With the extracted dense pixel semantic embedding~\cite{denseclip}, CAT-Seg~\cite{catseg} proposes to finetune CLIP with cost aggregation. EBSeg~\cite{ebseg} integrate Segment Anything Model~\cite{sam} into CLIP-based framework with image embedding balancing. \\
\noindent \textbf{Discussion with Previous Methods}\quad Our method MROVSeg is inspired by~\cite{san,fcclip}, but has significant differences: (1) Distinct with previous open-vocabulary segmentation methods, MROVSeg adapts multi-resolution CLIP features to open-vocabulary segmentation.  (2) The introduced Muti-grained Masked Attention scheme explicitly enforces the mask class recognition to aggregate both local and global semantics, which takes advantage of the internal consistency between multi-resolution features.

%% file: sec/sec_3_method.tex
\section{Method}
\label{sec:method}
In open vocabulary image segmentation task setting~\cite{simbase,ovseg}, an image $I\in \mathbb{R}^{3\times H \times W}$ is operated by a segmentation model $\mathcal{P}_{\theta}$ with parameter $\theta$ to produce a set of masks associated with $K$ categories:
\begin{equation}
    \{m_i,y_i\}_{i=1}^{K} = \mathcal{P}_{\theta}(I).
\end{equation}
The segmentation model $\mathcal{P}_{\theta}$ is trained on a base segmentation dataset (e.g., COCO~\cite{coco-stuff}) annotated with a fixed label set $\mathcal{Y}_{\mathrm{train}}=\{y_i\}_{i=1}^{K_{\mathrm{train}}}$ of $K_{\mathrm{train}}$ categories. And during test, model $\mathcal{P}_{\theta}$ is expected to segment objects of category set $\mathcal{Y}_{\mathrm{test}}$, which generally contains novel categories ($\mathcal{Y}_{\mathrm{test}} \not\subset \mathcal{Y}_{\mathrm{train}} $).

Accurate segmentation needs high-resolution image inputs. Due to low-resolution images used in vision-language pretraining, previous open-vocabulary methods employ extra image backbone (such as SAM~\cite{ebseg} and ResNet~\cite{scan}) to provide segmentation detail. Although recent studies~\cite{sed,fcclip} adapt convolution-based CLIP model for high-resolution training, but directly apply these methods to ViT-based CLIP model results in suboptimal performance~\cite{san} due to undesirable size adaptability of ViT. To this end, we propose MROVSeg, a ViT-based training framework to provide multi-resolution vision-language features for open-vocabulary image segmentation.
\subsection{Framework Overview}
\label{sec:framework_overview}
The overview of our training framework MROVSeg, for open-vocabulary image segmentation is shown in Fig. \ref{fig:overall_architecture}. At a high-level, an image is downsampled and padded to a low-resolution (such as $224\times224$) and processed by a pretrained CLIP ViT to extract global feature. To capture high-resolution local details, the high-resolution image is split into slices and input to the shared CLIP ViT encoder. These multi-resolution features are then concatenated with learnable queries and fed into a Multi-Res Adapter(Sec. \ref{sec:mrada}), to produce the fused features, query features, and attention masks used for mask predction(~\cref{mask_prediction}) mask classification(~\cref{mask_classification}). 

\subsection{Multi-Res Adapter}
\label{sec:mrada}
\begin{figure}[t]
    \centering
    \includegraphics[width=0.99\linewidth]{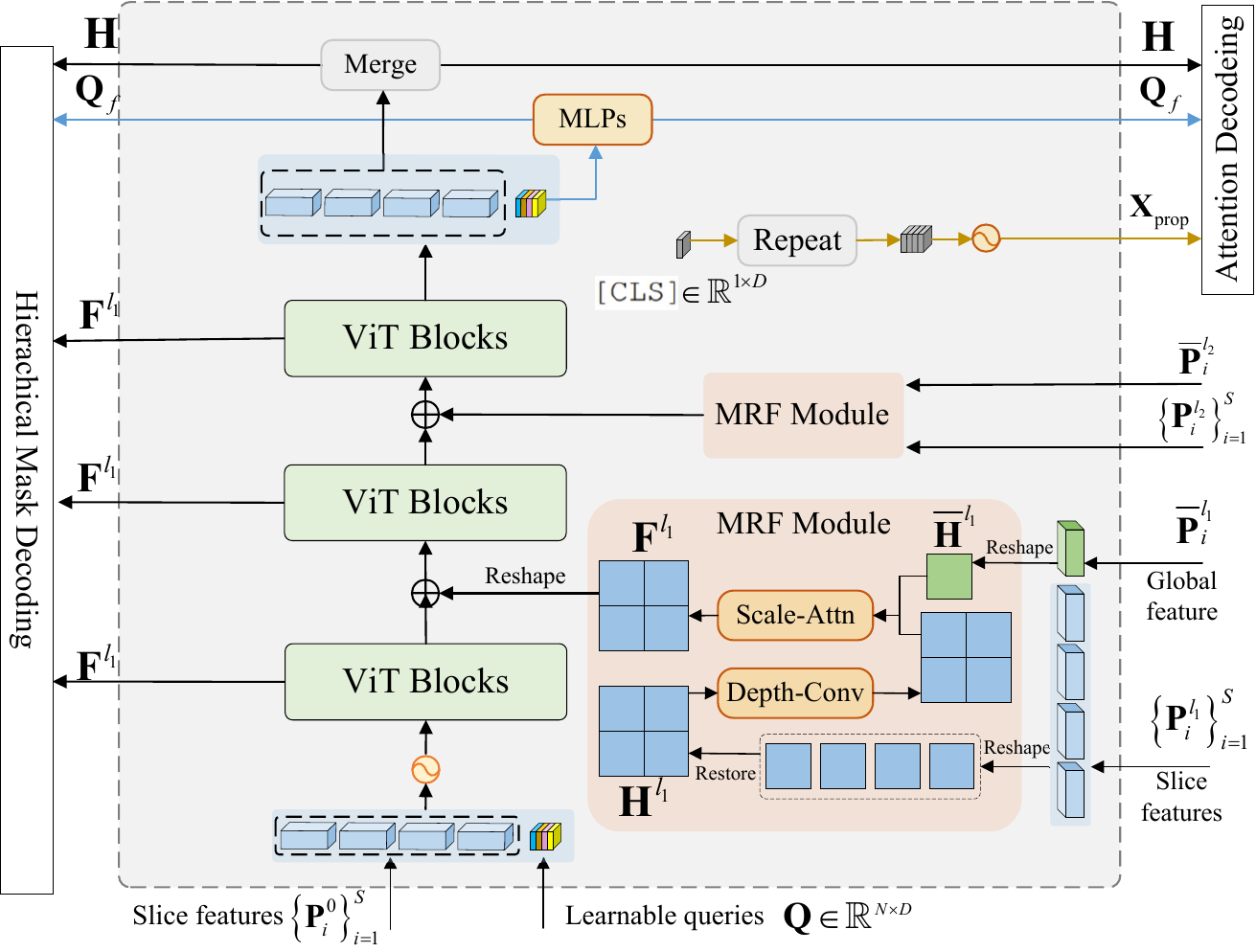}
    \caption{\textbf{Multi-Res Adapter.} The slice features from CLIP layer 0 $\{\mathbf{P}_i^0\}_{i=1}^{S}$ are concatenated with learnable queries and fed to ViT Blocks. The slice features from various CLIP layers are first adapted by MRF module to restore spatial geometry and capture long-range global contexts, then are injected to the intermediate ViT Blocks. The final output visual tokens and projected queries are utilized for downstream mask prediction and classification.}
    \label{fig: MR_Adapter} 
\end{figure}

As depicted in Fig. \ref{fig: MR_Adapter}, denote the slice (high resolution) features of $l$-layer as $\{\mathbf{P}_i^l\}_{i=1}^{S}$, the global feature (low resolution) as $\Bar{\mathbf{P}}_i^l$, where $\mathbf{P}_i \in \mathbb{R}^{L\times D}$, $S$ is the slice number, $L=H\times W$ is the token length, and $D$ is the channel number. Na\"ively concatenating the high resolution slice features for the subsequent segmentation prediction is promising, but have two defects: 1) the spatial geometry, such as the positional information, is corrupted across slice features $\{\mathbf{P}_i^l\}_{i=1}^{S}$; 2) the long-range dependencies and global context is missing. Thus, we propose Multi-Res Adapter to effectively restore spatial geometry of slice features and capture long-range context from global feature. In Multi-Res Adapter, the $0$-th slice features $\{\mathbf{P}_i^0\}_{i=1}^{S}$ are firstly concatenated with learnable queries $\mathbf{Q}\in \mathbb{R}^{N\times D}$ and input to the vanilla ViT blocks to build the query features for each objects. Then for target fusion layer $l$, the slice features $\{\mathbf{P}_i^{l}\}_{i=1}^{S}$ and global feature $\Bar{\mathbf{P}}_i^{l}$ are fused through a Multi-Res Fusion (MRF) Module and then injected into the ViT branch. \\
\noindent \textbf{Multi-Res Fusion (MRF) Module} first reshape the global feature to $\Bar{\mathbf{H}}^{l}\in \mathbb{R}^{H\times W\times D}$, and restore slice features to $\mathbf{H}^{l}\in \mathbb{R}^{m*H\times n*W \times D}$, where $S=m\times n$. To retain the spatial geometry of high-resolution features, we employ depth-wise separable convolutions to fuse the restored feature. To effectively model the local-global correspondence, we train a Scale-aware Attention~\cite{scaleattn} to fuse multi-res features into $\mathbf{F}^{l}\in \mathbb{R}^{H\times W\times D}$  as the fused feature
\begin{equation}
    \mathbf{F}^{l} = \texttt{Up}(a_{l})\odot\texttt{DConv}(\mathbf{H}^{l})+\texttt{Up}((1-a_{l})\odot \Bar{\mathbf{H}}^{l}).
\end{equation}
Then $\mathbf{F}^{l}$ is added to the visual tokens in Multi-Res Adapter. The scale attention decoder $f_{A}^{l}$ learns to predict the scale attention $a_{l}=\sigma(f_{A}^{l}(\mathbf{H}_{l}))\in [0,1]^{H\times W\times D}$ for layer $l$ to weigh the trustworthiness of low resolution context and high resolution detail. The sigmoid function $\sigma$ ensures $a_l$ weight in $[0,1]$, where $1$ means $a_l$ focus on high resolution detail. In practice, we empirically select the features from a CLIP layer set $\{l_j\}_{j=1}^{N_j}$ to apply in Multi-Res Adapter. For instance, for the model based on CLIP ViT-L model, $l\in\{\texttt{stem},6,12,18\}$. The fused features $\{\textbf{F}^{l_j}\}_{j=1}^{N_j}$ are used for hierarchical mask decoding. The final layer output slice features $\{\mathbf{P}_i^0\}_{i=1}^{S}$ are restored to $\mathbf{H}\in\mathbb{R}^{m*H\times n*W\times D}$ as the visual feature for hierarchical mask decoding and multi-grained masked attention. And the output queries $\mathbf{Q}$ are projected as the query feature
\begin{equation}
    \mathbf{Q}_{f} = \texttt{MLP}^{\mathrm{Q}}(\mathbf{Q}), 
\end{equation}
for hierarchical mask decoding and multi-grained masked attention. 
\subsection{Mask Prediction} 
\label{mask_prediction}
\textbf{Hierarchical Mask Decoding}\quad High-resolution features preserve more spatial detail, thus benefit segmentation, especially for mask prediction~\cite{detr}. However, directly upsampling features is computationally demanding. Thus, similar to FPN, we first upsample the multi-resolution features $\{\mathbf{F}^{l_j}\}_{j=1}^{3}$ from the Multi-Res Adapter by $\{2\times, 4\times, 8\times\}$ to build the feature pyramid. Then we gradually concatenate the multi-resolution features with the final visual feature $\mathrm{\mathbf{H}}$ at channel dimension and upsample by 2 transposed convolution layers $\mathrm{\mathbf{H}}_{\texttt{up}} = \texttt{TransposeConv}(\texttt{cat}(\mathrm{ \mathbf{H}},\mathbf{F}^{l_j}))$. Finally, we project the upsampled feature $\mathrm{\mathbf{H}}_{\texttt{up}}$ to the pixel feature space by $\texttt{MLP}$ then decode the mask by inner product of query feature and mask feature
\begin{equation}
    \mathrm{\mathbf{H}}_{\texttt{pix}} = \texttt{MLP}^{\texttt{pix}}(\mathrm{\mathbf{H}}_{\texttt{up}}),
\end{equation}
\begin{equation}
    \mathrm{\mathbf{M}}_{\texttt{mask}} = \mathrm{\mathbf{Q}}_f \times \mathrm{\mathbf{H}}_{\texttt{pix}},
\end{equation}
where the query feature $\mathrm{\mathbf{Q}}_f$ is from the Multi-Res Adapter which described in Sec. 4.3. $\mathrm{\mathbf{M}}_{\texttt{mask}}\in [0,1]^{N\times H \times W}$ is the mask prediction, and we omit the sigmoid function in Eq.5.

\subsection{Mask Classification}
\label{mask_classification}
\begin{figure}[t]
    \centering
    \includegraphics[width=0.99\linewidth]{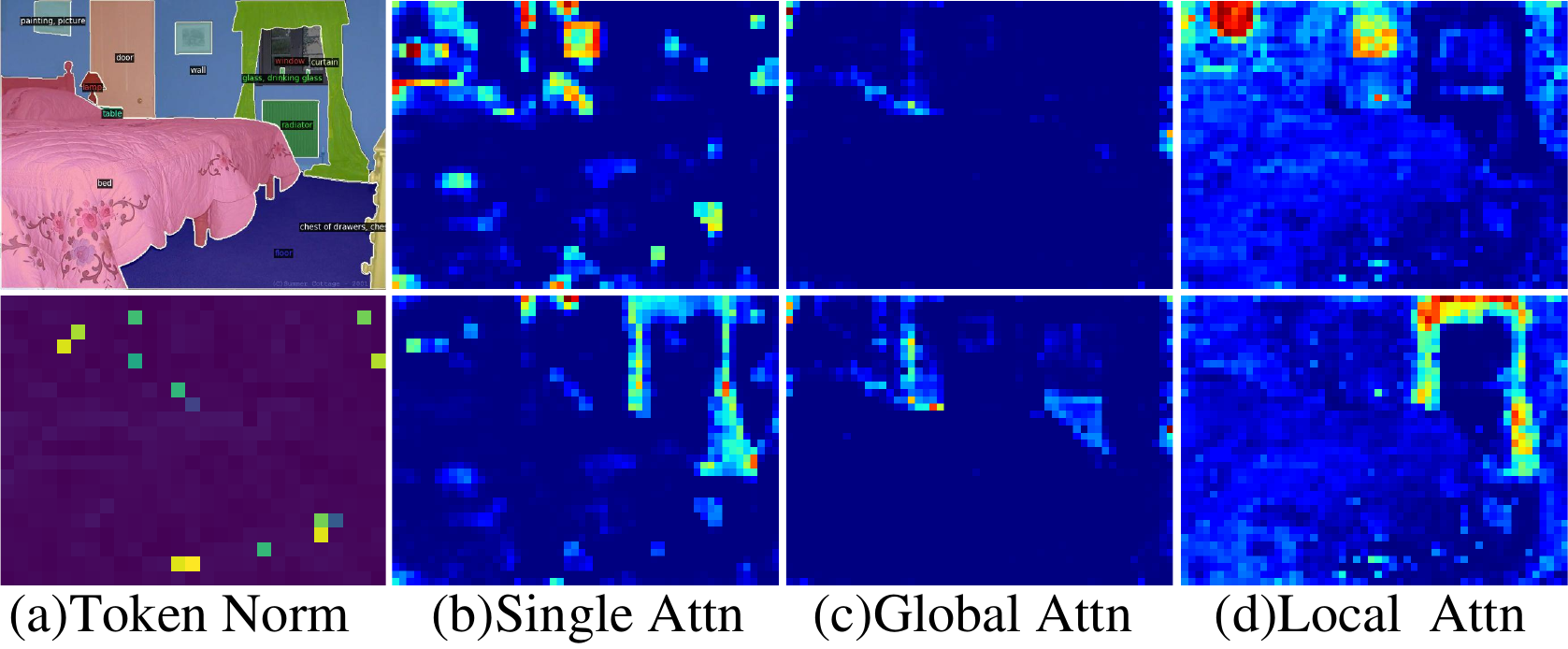}
    \caption{\textbf{Effect of decoupled attention decoding for multi-grained semantics.} With single attention mask decoding, the spatial cues are overwhelmed by background noise (b). Our decoupled attention mask decoding effectively splits the global and local semantics, producing relatively clean global (c) and local (d) attention masks.}
    \label{fig: decoupattn}
\end{figure}
\begin{figure}
    \centering
    \includegraphics[width=0.9\linewidth]{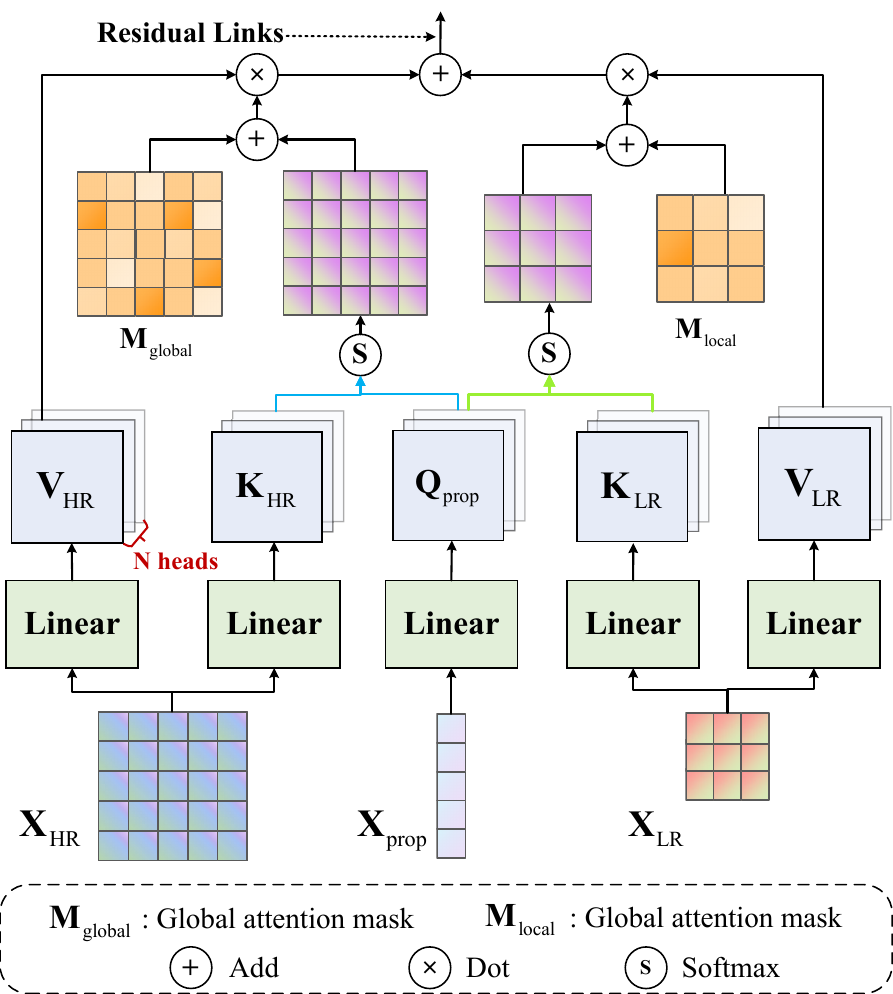}
    \caption{\textbf{Multi-grained Masked Attention.} Object \texttt{[CLS]} tokens $\mathrm{\mathbf{X}}_{\texttt{prop}}$ perform cross attention with high- and low-resolution CLIP features $\mathbf{X}_{\texttt{LR}}$ and $\mathbf{X}_{\texttt{HR}}$  with decoupled attention masks.}
    \label{fig:multi-grained} 
\end{figure}

Recent ViT-based open-vocabulary segmentation methods~\cite{maskclip,san,ebseg} perform mask class recognition by predicting attention masks to guide the attention maps of original CLIP \texttt{[CLS]} token on the region of interests in the intermediate layers. We observe that the predicted attention masks (shown in~\cref{fig: decoupattn}(b)) in these methods tend to be overwhelmed by heavy background noises which spatially related to \textbf{high norm} tokens (shown in~\cref{fig: decoupattn}(a)), leading to the unsatisfactory classification performance. While prior ViT pretraining technique~\cite{vitreg} indicates these high norm tokens contain rich global contexts, recent advances in training-free open-vocabulary segmentation~\cite{wang2025sclip,shao2025explore,lan2024proxyclip} reveal that high norm tokens can easily disturb spatial correlation in CLIP features. Inspired by these inspection, we hypothesize that CLIP features hold internal consistency among multi-resolution inputs, and propose to simultaneously aggregate semantics from multi-resolution CLIP features by predicting decoupled attention masks. \\
\noindent \textbf{Decoupled Attention Mask Decoding}\quad To sufficiently aggregate multi-grained semantics from CLIP, we first duplicate the \texttt{[CLS]} token to query number $N$ and create learnable positional embedding for them, dubbed as the $\mathrm{\mathbf{X}_{\texttt{prop}}}\in \mathbb{R}^{N\times D}$. We aim to enforce $\mathrm{\mathbf{X}_{\texttt{prop}}}$ to extract the global image-wise and local object-specific semantics from low- and high-resolution CLIP feature respectively. Thus, for visual feature $\mathbf{H}$, we first extract global contexts with max pooling $\bar{\mathbf{H}}=\texttt{MaxPool}(\mathbf{H})$ and train \texttt{MLP}s project them to attention space
\begin{equation}
    \mathbf{A}_{\mathrm{local}} = \texttt{MLP}^{\texttt{L}}(\mathbf{H}),\quad\mathbf{A}_{\mathrm{global}}= \texttt{MLP}^{\texttt{G}}(\bar{\mathbf{H}}),
\end{equation}
where $\mathbf{A}_{\texttt{global}}\in\mathbb{R}^{H \times W \times D'},\mathbf{A}_{\texttt{local}}\in\mathbb{R}^{m*H \times n*W \times D'}$ denote the local and global attention features respectively. Then we decode local and global per-head attention masks by the inner product with 
\begin{equation}
    \mathbf{M}_{\mathrm{local}} = \mathbf{Q}_{f} \times \mathbf{A}_{\mathrm{local}}^{\mathrm{T}},\enspace \mathbf{M}_{\mathrm{global}} = \mathbf{Q}_{f} \times \mathbf{A}_{\mathrm{global}}^{\mathrm{T}},
\end{equation}
where $\mathbf{Q}_{f}$ is the output query feature described in Sec.\ref{sec:mrada}. We show this decoupled \textit{resolution-aware} attention decoding benefit the multi-grained aggregation in Fig.\ref{fig: decoupattn}.\\ 
\noindent\textbf{Multi-grained Masked Attention}\quad As shown in Fig.\ref{fig:multi-grained}, we perform cross attention to update the $\mathrm{\mathbf{X}}_{\texttt{prop}}$ with multi-resolution CLIP features, with the predicted attention masks $\mathbf{M}_{\mathrm{local}}$ and $\mathbf{M}_{\mathrm{global}}$,
\begin{equation}
    \!\!\mathbf{X}^{l+1}_{\texttt{prop}}\! =\! \texttt{softmax}(\mathbf{Q}^{l}_{\texttt{prop}}\!\!\left[\arraycolsep=1.0pt\def\arraystretch{1.5}\begin{array}{c}\mathbf{K}^{l}_{\texttt{LR}}  \\ \mathbf{K}^{l}_{\texttt{HR}}\\\end{array}\right]^{\!\mathrm{T}}\!\!\!\!+\!\notag\left[\arraycolsep=1.0pt\def\arraystretch{1.5}\begin{array}{c}   \mathbf{M}_{\mathrm{global}} \\\mathbf{M}_{\mathrm{local}}\end{array}\right]^{\!\mathrm{T}}\!\!\!)\!\left[\arraycolsep=1.0pt\def\arraystretch{1.5}\begin{array}{c}\mathbf{V}^{l}_{\texttt{{LR}}}  \\ \mathbf{V}^{l}_{\texttt{{HR}}}\end{array}\right]\! +\mathbf{X}^{l}_{\texttt{prop}} , \tag{8}
\end{equation}
where $\mathbf{Q}_{\texttt{prop}} = \mathbf{W}_{\mathrm{q}}^\mathrm{T}\mathbf{X}_{\texttt{prop}}$ is query embeddings. Denote the low- and high-resolution CLIP tokens as $\mathbf{X}_{\texttt{LR}}$ and $\mathbf{X}_{\texttt{HR}}$. $\mathbf{K}_{\texttt{LR}} = \mathbf{W}_{\mathrm{k}}^\mathrm{T}\mathbf{X}_{\texttt{LR}}$ and $\mathbf{K}_{\texttt{HR}} = \mathbf{W}_{\mathrm{k}}^\mathrm{T}\mathbf{X}_{\texttt{HR}}$ are the key embeddings of low- and high-resolution CLIP visual tokens respectively. $\mathbf{V}_{\texttt{LR}} = \mathbf{W}_{\mathrm{v}}^\mathrm{T}\mathbf{X}_{\texttt{LR}}$ and $\mathbf{V}_{\texttt{HR}} = \mathbf{W}_{\mathrm{v}}^\mathrm{T}\mathbf{X}_{\texttt{HR}}$ are value embeddings. $\mathbf{W}_{\mathrm{q}}$,$\mathbf{W}_{\mathrm{k}}$ and $\mathbf{W}_{\mathrm{v}}$ are projection weights of cross-attention layer. The final output proposal logits $\mathbf{X}_{\texttt{prop}}$ are projected to the shared vision-language space and compute cosine similarity with text embeddings to obtain proposal logits $\mathbf{C} \in \mathbb{R}^{N\times K}$: $\mathbf{C}=\mathbf{X}_{\texttt{prop}}\mathbf{W}_{\mathrm{visual}}\mathbf{W}_{\mathrm{text}}^{\mathrm{T}}\mathbf{X}_{\texttt{text}}^{\mathrm{T}}$, where $K$ is the number of categories, and $\mathbf{W}_{\mathrm{visual}}$ and $\mathbf{W}_{\mathrm{text}}$ are projection weights. Finally, the final segmentation map $\mathbf{S}\in\mathbb{Z}_{+}^{K\times H \times W}$ is produced by 
\begin{equation}
    \mathbf{S} = \mathbf{C} \times \mathbf{M}_{\texttt{mask}}^{\mathrm{T}}. \tag{9}
\end{equation}
\noindent
\textbf{Image-conditioned Text Feature}\quad Recent studies~\cite{maftp,catseg} reveal that CLIP text encoder struggles to generate discriminative text embeddings for similar categories. Thus, we follow MAFT-Plus~\cite{maftp} to condition the text embeddings with learnable cross attention layers between text embeddings and regional pooling visual features: $\mathbf{X}_{\texttt{text}} = \texttt{CrossAttnLayer}(\mathbf{X}_{\texttt{text}}^{ori}, \texttt{MaxPool}(\mathbf{H}))$, where $\mathbf{H}$ is depicted in \cref{sec:mrada}. $\mathbf{X}_{\texttt{text}}^{ori}$ is the original text embedding extract by CLIP text encoder.

%% file: sec/sec_4_experiments.tex
\section{Experiments}
\input{tables/main_results}
\subsection{Settings}
For open-vocabulary semantic segmentation task, We train our models on COCO-Stuff~\cite{coco-stuff} dataset which comprises 164K images with densely annotated masks spanning 171 categories. Then we evaluate MROVSeg on five well-established open-vocabulary semantic segmentation benchmarks for standard evaluation. We further evaluate MROVSeg on Cityscapes~\cite{cordts2016cityscapes} benchmarks to explore the ability of handling high-resolution image input. We follow common practice~\cite{simbase,zs3net} to measure the segmentation performance by mean intersection over union (mIoU) score. For open-vocabulary panoptic segmentation task, we train MROVSeg on COCO-Panoptic~\cite{coco} dataset. Then we evaluate zero-shot performance of MROVSeg on ADE~\cite{ade} panoptic benchmark, and measure the panoptic segmentation performance by  panoptic quality (PQ), segmentation qualitiy (SQ) and recognition quality (RQ).\\
\noindent \textbf{Datasets}\quad 
The standard semantic segmentation benchmarks contains three dataset: ADE~\cite{ade}, Pascal Context~\cite{pc}, and Pascal VOC~\cite{voc}. The ADE dataset contains around 20K and 2K images for training and validation, respectively. This dataset is annotated with 150 and 847 categories, resulting in two separate segmentation benchmarks, namely ADE-150 and ADE-847. Similarly, the Pascal Context dataset has 5K images for both training and validation. It is annotated with 59 and 459 classes, forming two benchmarks known as PC-59 and PC-459. The Pascal VOC dataset comprises 1464 and 1449 images for training and validation, encompassing annotated masks across 20 semantic categories. \\
\input{tables/panopticseg}\\
\noindent\textbf{Implementation Details}\quad We adopt the vanilla ViT block as transformer block in Multi-Res Adapter, and we use 6 blocks with 12 attention heads, 100 query tokens, feature dimension is 768 as default. We choose OpenAI pretrained ViT-based CLIP~\cite{clip} models in all experiments for better reproducibility. We empirically choose $\texttt{[CLS]}$ token from CLIP layer 9, CLIP ViT-B/16 model, and subsequent layers for Multi-grained Masked Attention. i.e., we use last 3 blocks for Multi-grained Masked Attention. For CLIP ViT-L/14 model, we $\texttt{[CLS]}$ token from CLIP layer 18, and use subsequent layers for Multi-grained Masked Attention. Our models are trained with image input resolution $640\times 640$ , and slice and downsample to $320\times 320$ to fit the CLIP input resolution. More implementation details are in Supplementary Material.\\
\begin{figure}[tb]
  \centering
    \begin{minipage}{0.58\linewidth}
     \centering
     \includegraphics[width=\linewidth]{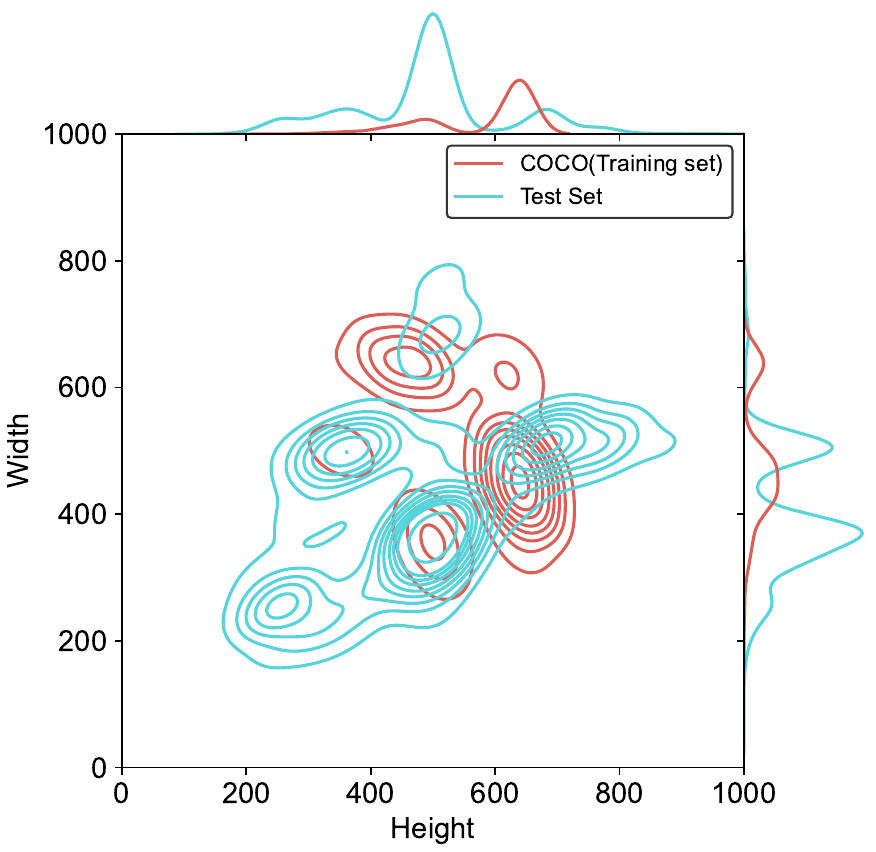}
     \caption{Visualization of image resolution distribution histogram of the datasets in Tab.1.}\label{fig:res distribution}
   \end{minipage}\hfill
  \begin{minipage}{0.35\linewidth}
     \centering
     \includegraphics[width=\linewidth]{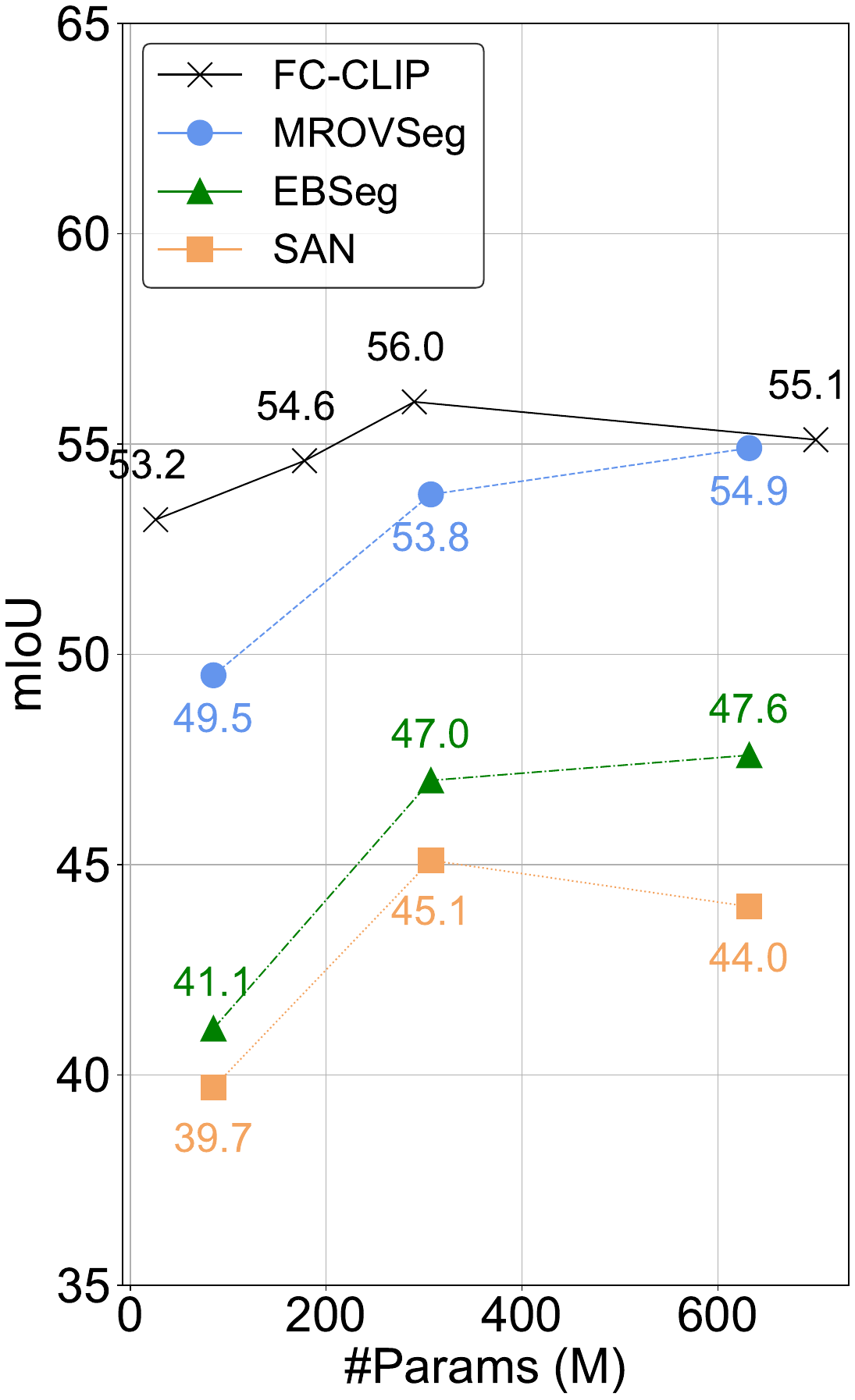}
     \caption{Effect of scaling up backbone on Cityscapes.}\label{fig:scalability}
   \end{minipage}\hfill
  \vspace{-20pt}
\end{figure}
\vspace{-20pt}

\subsection{Accuracy Evaluation}
\begin{figure}[t]
  \centering
    \includegraphics[width=0.99\linewidth]{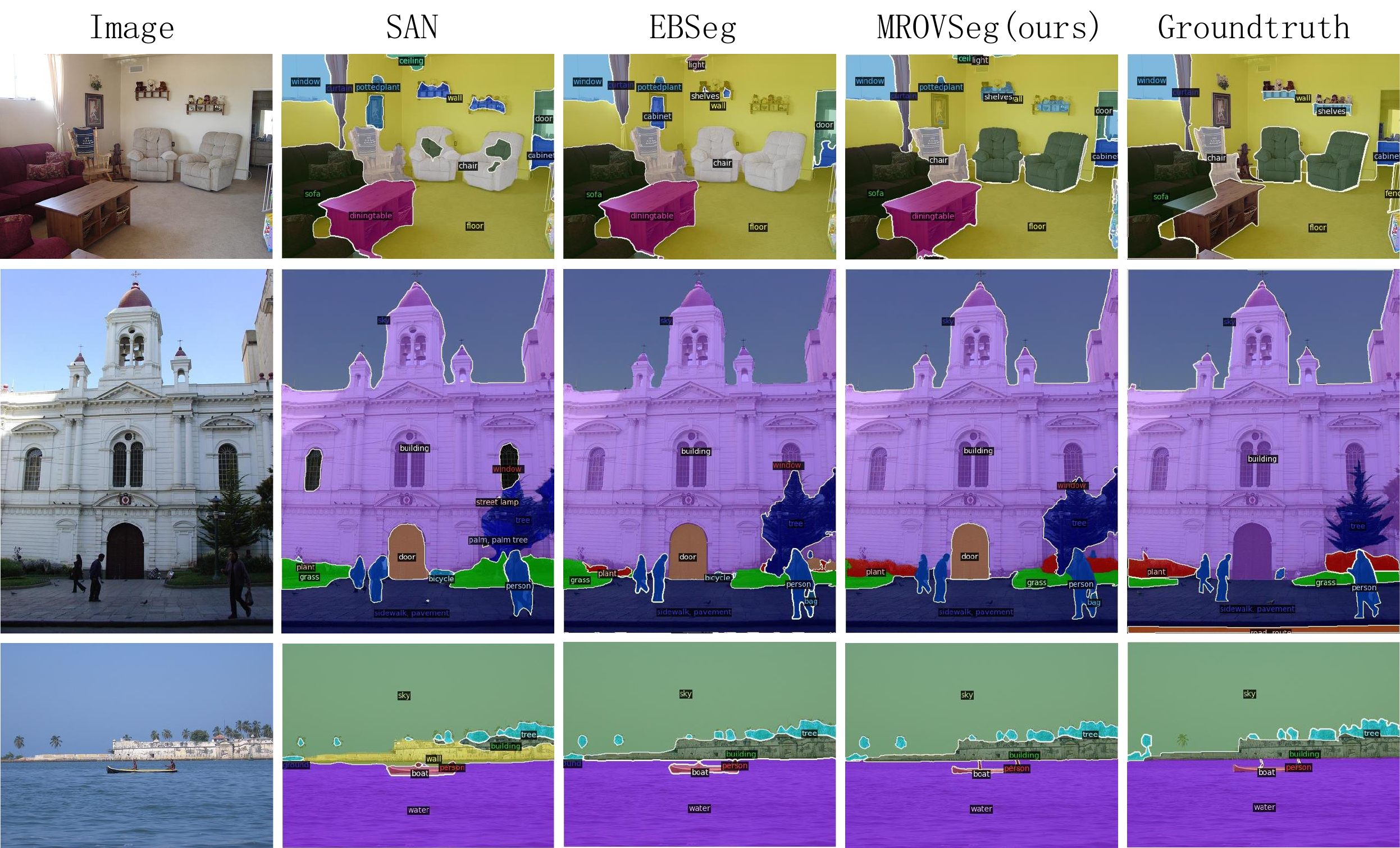}
    \caption{Qualitative comparison with SAN~\cite{san} and EBSeg~\cite{ebseg}.}
    \label{fig:qualitative_comparison1}
  \hfill
  \vspace{-15pt}
\end{figure}

\textbf{Open-Vocabulary Semantic Segmentation}\quad We compare the semantic segmentation performance of MROVSeg with current state-of-the-art methods in~\cref{tab:main_res}. First of all, our method significantly outperforms other state-of-the-art methods all various on most open-vocabulary semantic segmentation benchmarks. Specifically, our method supasses state-of-the-art method CAT-Seg~\cite{ebseg} with the same CLIP backbones on four benchmarks by remarkable margins ($+0.9 \%$ mIoU for ADE-847,  $+0.2 \%$ mIoU for PC-459, $+0.6 \%$ mIoU for ADE-150, $+1.2 \%$ mIoU for PC-59,  and $+1.2$ for VOC-20 with CLIP ViT-B backbone, $+0.4 \%$ mIoU for ADE-847,  $+0.2 \%$ mIoU for PC-459,  $+1.0 \%$ mIoU for PC-59,  and $+0.6$ for VOC-20 with CLIP ViT-B backbone). Compared to methods with additional image backbones, our model outperforms EBSeg~\cite{ebseg}(with SAM) by $+2.7, +3.0, +4.1, + 4.1, + 1.4$ mIoU\% for ADE-847, PC-459, ADE-150, PC-59, and VOC-20 respectively. In addition, our models outperforms other convolution-based high-resolution trained method FC-CLIP~\cite{fcclip} and SED~\cite{sed}. \cref{fig:qualitative_comparison1} shows the qualitative comparison between MROVSeg and state-of-the-art methods (SAN~\cite{san} and EBSeg~\cite{ebseg}). Evidently, MROVSeg can segment objects more precisely (the first row, class \textit{sofa}), and provide more detailed mask prediction (the second and third row, more accurate object boundaries). 
\par The size adaptability of ViTs have been demonstrated~\cite{fcclip} to be worse than ConvNets. \cref{fig:res distribution} shows that the image sizes of the datasets evaluated in \cref{tab:main_res} are primarily distributed from $[250,800]$. To evaluate the ability of handling high-resolution image input, we retrain MROVSeg on COCO-Stuff with $1024\times1024$ resolution evaluate performance on Cityscapes benchmark. Notably, MROVSeg reaches comparable performance to convolution-based methods~\cite{fcclip} while scaling up the backbone, which greatly enhanced than other ViT-based methods (EBSeg~\cite{ebseg} and SAN~\cite{san}). 
\input{tables/ratiop}
\input{tables/efficiency}\\
\noindent \textbf{Open-Vocabulary Panoptic Segmentation}\quad In \cref{tab:panoptic}, We evalute the panoptic segmentation performance of MROVSeg on mainstream open-vocabulary panoptic segmentation benchmarks ADE~\cite{ade}. It is noteworthy that our method surpass previous arts on panoptic quality (PQ) and recognition quality (RQ), achieving new state-of-art open-vocabulary panoptic segmentation performance. Furthermore, close-set panoptic segmentation results indicate that training MROVSeg model does not lead to severe overfitting issue on the base dataset.


\vspace{-5pt}
\subsection{Efficiency Evaluation}
\vspace{-5pt}
\noindent
\textbf{Memory Consumption}\quad Denote the slice ratio as $p=\frac{slice\;\,resolution}{input\;\,resolution}$,  we examine the effect of $p$ value in Tab.~\ref{tab:crop ratio}. Notably, only using single resolution feature, i.e., $p=0$, or directly adopting high-resolution image as CLIP input, i.e., $p=1.0$ lead to significant performance degradation. While some $p$ values with overlapped slicing (such as $0.625$) obtains better performance on some datasets(such as ADE-150 and VOC-20), we choose the default $p$ value as $0.5$ considering computation overhead.\\
\noindent \textbf{Computation Overhead}\quad 
We compare the computation overhead of MROVSeg with recent methods~\cite{maskclip,ovseg,catseg,ebseg,fcclip} in Tab.~\ref{tab:efficiency}. We measure the number of parameter, giga floating-point operations per second, inference FPS and training time. Our method exhibits strong efficiency among these methods in terms of both training and inference. This is achieved by 1) single CLIP backbone do not need additional image backbones; 2) slice then input strategy avoid quadratic computation cost with regard to input image size. More detailed parameter settings are in Supplementary Material.
\input{tables/ab_componets}
\input{tables/incremantal_comp}
\subsection{Ablation Study}
\input{tables/ab_mradapter}
\label{ablationstudy}
For ablation experiments, except for \cref{tab:conv_counterparts}, we all employ our CLIP ViT-B/16 based model as our ablated baseline.\\
\noindent \textbf{Components Analysis}\quad Tab.~\ref{tab:components} shows the performance effect of the main components in MROVSeg. The baseline adopts single resolution CLIP plain ViT feature for mask decoding, and uses the MaskPooling~\cite{odise} for mask class recognition. The baseline obtains the mIoU scores of 52.3\%, 22.4\%, 8.5\%, 10.8\% and 93.6\% for PC-59, ADE-150, ADE-847, PC-459 and VOC-20. We first introduce multi-resolution CLIP features, which marginally outperforms baseline. Then we introduce the Multi-Res Adapter, the model significantly outperforms baseline by 4.8\%, 5.1\%, 2.3\%, 7.2\% and 1.5\%. Next we integrate the Masked-Attention~\cite{san} to the model, it slightly outperforms MaskPooling. Finally we integrate Multi-grained Masked Attention, the model performance reaches 58.7\%, 32.4\%, 12.9\%, 19.2\% and 95.8\% for PC-59, ADE-150, ADE-847, PC-459 and VOC-20, which outperforms baseline by 6.4\%, 7.0\%, 3.9\%, 8.4\% and 2.2\%. Futhermore, we show the effect of of Hierarchical Mask Decoding and Image-conditioned Text Feature in \cref{tab:inremental}, both showing consistent improvements.\\
\noindent \textbf{Effect of Multi-Res Adapter}\quad We first conduct experiment to examine the different micro-design choice in Multi-Res Adapter in Tab.~\ref{fig: MR_Adapter}. In (a), we examine different spatial restore strategies in MRF module. The depthwise convolution is significantly better than concatenation. And we present the impact of local-global modeling methods (b), and the best results on all benchmarks are achieved by scale attention. Then for the ViT blocks setting, we present the effect of block number in (c). Increasing the block number to 9 brings limited improvement while incurring heavy computation cost. In (d), we examine the effect of CLIP layers whose feature we adopt. In (e), we present the impact of channel width we use. Finally, we examine the effect of query number in (f). We further compare the performance of MROVSeg with ViT-based and convolution-based CLIP models in \cref{tab:conv_counterparts}. The results indicate that ViT-based encoder with a sliding window approach outperforms convolution-based counterpart.
\begin{figure}[t]
    \centering
    \includegraphics[width=0.99\linewidth]{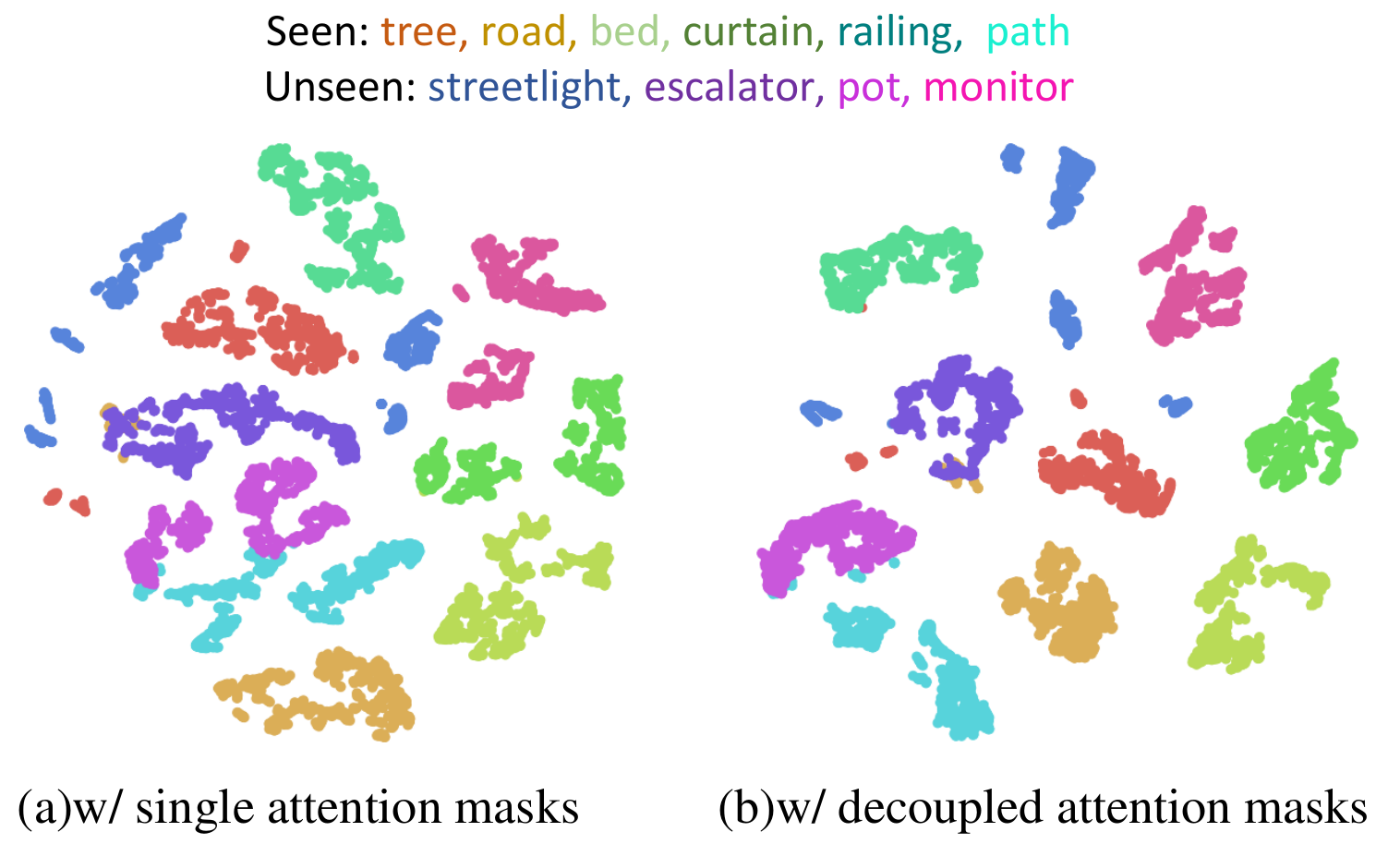}
    \vspace{-10pt}
    \caption{\textbf{Effects of decoupled attention decoding for multi-grained semantics.} The t-SNE~\cite{van2008visualizing} visualization shows that decoupled attention masks can further split the query embedding from different classes, making clear classification boundaries.}
    \label{fig: tsne}
\end{figure} 
\input{tables/convolution_conterpart}\\
\noindent \textbf{Effect of Multi-grained Masked Attention}\quad As mentioned before, \cref{tab:components} and \cref{fig: decoupattn} quantitatively and qualitatively show the effectiveness of Multi-grained Masked Attention respectively. We further visualize the query \texttt{[CLS]} embeddings (
\ie, $\mathbf{X}_{\texttt{prop}}$) by t-SNE~\cite{van2008visualizing} dimensionality reduction within ADE-150~\cite{ade} benchmarks in \cref{fig: tsne}. We color the embeddings based on the Hungarian matching with groundtruth. In (a), we can observe that the queries from the same classes are well-posed as masked attention can aggregate semantics. Conversely, with decoupled attention masks, the query embeddings from different classes are split further in (b), indicating the effectiveness of multi-grained masked attention in mask classification.\\

%% file: tables/main_results.tex
\definecolor{light-gray}{gray}{0.9}
\begin{table*}[t]
\scriptsize
\centering
\caption{\textbf{Semantic segmentation performance comparison with state-of-the-art methods.} \dag:~We cite the reproduction performance of these methods\cite{simbase,zegformer} trained with full COCO Stuff dataset from previous works\cite{san, catseg}. PC: Pascal Context, VOC: Pascal VOC.}
\resizebox{1.0\textwidth}{!}{
\begin{tabular}{c|cccc|ccccc}
\toprule
\multirow{1}{*}{Method} & 
\multirow{1}{*}{VL-Model} & 
\multirow{1}{*}{Training Dataset} &
End to End Training &
Extra Backbone &
ADE-847   & 
PC-459    & 
ADE-150   & 
PC-59     & 
VOC-20       
\\ \hline\hline

Zegformer\dag~\cite{zegformer}&
CLIP ViT-B/16 &
COCO Stuff &
\XSolidBrush &
ResNet-101 &
5.6 &
10.4 &
18.0 &
45.5 &
89.5 
\\

ZSSeg\dag~\cite{simbase}   & 
CLIP ViT-B/16             & 
COCO Stuff                         &     
\XSolidBrush &
ResNet-101 &
6.9                                     & 
9.7                                     & 
21.1                                    & 
51.9                                   & 
91.8            
\\

OvSeg~\cite{ovseg}        & 
CLIP ViT-B/16             & 
COCO Stuff                               & \XSolidBrush &
ResNet-101c &
7.1                                     &
11.0                                    &
24.8                                    & 
53.3                                   &
92.6            

\\

SAN~\cite{san}         & 
CLIP ViT-B/16            & 
COCO   Stuff        &   
\Checkmark &
- &
10.7   & 
13.7       & 
28.9        &  
55.4          &   
94.6                
\\

EBSeg~\cite{ebseg}          & 
CLIP ViT-B/16            & 
COCO   Stuff        &     
\XSolidBrush &
SAM-B &
11.1   & 
17.3       & 
30.0        &  
56.7          &   
94.6                      
\\

SCAN~\cite{scan}          & 
CLIP ViT-B/16            & 
COCO   Stuff        &     
\XSolidBrush &
ResNet-101 &
10.8   & 
13.2       & 
30.8        &  
\underline{58.4}          &   
\textbf{97.0}                      
\\

SED~\cite{sed}         & 
CLIP ConvNeXt-B            & 
COCO   Stuff        &     
\Checkmark &
- &
11.2   & 
18.6       & 
31.6        &  
57.3          &   
94.4                    
\\

CAT-Seg~\cite{catseg} &
CLIP ViT-B/16 &
COCO Stuff &
\Checkmark &
-&
\underline{12.0} &
\underline{19.0} &
\underline{31.8} &
57.5 &
94.6 
\\

\rowcolor{light-gray}
MROVSeg &
CLIP ViT-B/16 &
COCO Stuff &
\Checkmark &
- &
\textbf{12.9} &
\textbf{19.2} &
\textbf{32.4} &
\textbf{58.7} &
\underline{95.8}
\\ \hline

OvSeg~\cite{ovseg}         & 
CLIP ViT-L/14             &
COCO Stuff                             &  
\XSolidBrush &
Swin-B &
9.0                                     &
12.4                                    & 
29.6                                    & 
55.7                                   & 
94.5              
\\

MaskCLIP~\cite{maskclip}       & 
CLIP ViT-L/14             &
COCO Panoptic             &  
\XSolidBrush &
ResNet-50 &
8.2                                     &
10.0                                    &
23.7                                    & 
45.9                                   & 
-         
\\

ZSSeg\dag~\cite{simbase}     & 
CLIP ViT-L/14             & 
COCO Stuff                        &   
\XSolidBrush &
ResNet-101 &
7.1                                     &
10.2                                    & 
21.7                                    & 
52.2                                   & 
92.3                     
\\

SAN~\cite{san}     &
CLIP ViT-L/14         & 
COCO Stuff        &   
\Checkmark &
- &
13.7                        & 
17.1                       &
33.3                        & 
60.2                       & 
95.5                
\\ 

ODISE~\cite{odise} &
CLIP ViT-L/14 &
COCO Panoptic &
\XSolidBrush &
Stable Diffusion &
11.0&
13.8&
28.7&
55.3&
-  
\\

EBSeg~\cite{ebseg}          & 
CLIP ViT-L/14            & 
COCO   Stuff        &     
\Checkmark &
SAM-B &
13.7   & 
21.0       & 
32.8        &  
60.2          &   
96.4              
\\

SCAN~\cite{scan}         & 
CLIP ViT-L/14            & 
COCO   Stuff        &   
\XSolidBrush  &
ResNet-101 &
14.0   & 
16.7       & 
33.5        &  
59.3          &   
97.2               
\\

FC-CLIP~\cite{fcclip} &
CLIP ConvNeXt-L &
COCO Panoptic &
\XSolidBrush &
- &
14.8 &
18.2 &
34.1 &
58.4 &
95.4 
\\

SED~\cite{sed}          & 
CLIP ConvNeXt-L            & 
COCO   Stuff        &  
\Checkmark &
- &
13.7   & 
22.1       & 
35.3        &  
60.9          &   
96.1                 
\\

CAT-Seg~\cite{catseg} &
CLIP ViT-L/14 &
COCO Stuff &
\Checkmark &
- &
\underline{16.0} &
\underline{23.8} &
\textbf{37.9} &
\underline{63.3} &
\underline{97.0} 
\\

MAFT+~\cite{maftp} &
CLIP ConvNext-L &
COCO Stuff &
\Checkmark &
- &
15.1 &
21.6 &
36.1 &
59.4 &
96.5 
\\

\rowcolor{light-gray}
MROVSeg&
CLIP ViT-L/14 &
COCO Stuff &
\Checkmark &
- &
\textbf{16.4} &
\textbf{24.0} &
\underline{36.9} &
\textbf{64.3} &
\textbf{97.6} 
\\
\bottomrule
\end{tabular}
}
\label{tab:main_res}
\end{table*}

%% file: tables/panopticseg.tex
\begin{table}[t]
    \centering
    \caption{\textbf{Panoptic segmentation performance comparison with state-of-the-art methods.} }
    \scriptsize
    \begin{tabular}{c|ccc|ccc}
    \toprule
      \multirow{2}{*}{}  & & ADE & & & COCO &  \\
       &    \multicolumn{3}{c|}{open-vocabulary}  & \multicolumn{3}{c}{close-set}  \\
       Method & PQ &  SQ &  RQ  & PQ & SQ & RQ \\
        \hline \hline
        FreeSeg~\cite{qin2023freeseg} & 16.3 &  - &  -  & - & -    &     \\
        MaskCLIP~\cite{maskclip} & 15.1 &  70.4 &  19.2  & - & -    & -    \\
       ODISE~\cite{odise} & 22.6 &  - &  -  & - & -    &     \\
       OPSNet~\cite{opsnet} & 19.0 &  52.4 &  23.0  & - & -  & -       \\
        FCCLIP~\cite{fcclip}    & 26.8 & 71.5 & 32.2 & 54.4 & 44.6  & 63.7 \\
       MAFT-Plus~\cite{maftp} & 27.1 &  \textbf{73.5}  & 32.9 & - & - & - \\
        MROVSeg & \textbf{27.3}   &  72.8 & \textbf{33.4} & 52.0 & 41.4  & 60.9  \\
      \bottomrule
    \end{tabular}
    \vspace{-10pt}
    \label{tab:panoptic}
\end{table}

%% file: tables/ratiop.tex
\begin{table}[tb]
\centering
\caption{\textbf{Effect of different crop ratio $p$.} We also report \#GFLOPS and GPU memory consumption (MB). For $p>0.5$, we adopt overlapped slicing. }
\vspace{-5pt}
\label{tab:crop ratio}
\setlength{\tabcolsep}{4.4pt}
\scriptsize
    \begin{tabular}{cccccccc}
    \toprule
         $p$        & Mem. & GFLOPS  &   A-150  & PC-59 & A-847 & PC-459 & VOC-20  \\ 
        \hline 
        $0$ & 6768 &   116.5   & 28.6  &  54.7  & 10.6 & 16.4 & 94.5 \\
        $1.0$ & 13493 & 293.1  & 28.8  & 54.5 & 10.5 & 15.6 & 94.4\\
      $0.75$ & 16261 &  225.7   &   31.5   &   54.0  & 11.5 & 15.9  & 95.1 \\
      $0.625$ & 12942 &  184.5 &  32.1   &  \textbf{59.1}   &  12.0 & \textbf{19.6} & 95.6\\
      \underline{$0.5$}  &9779 & 142.8 & \textbf{32.4} &  58.7   & \textbf{12.9}  & 19.2 & \textbf{95.8}       \\
      $0.25$ & 7560& 92.5 & 27.5 &  55.7   & 9.9 &  16.0 & 93.7    \\
      \bottomrule
    \end{tabular}
    \vspace{-10pt}
\end{table}

%% file: tables/efficiency.tex
\begin{table}[t!]
\centering
\caption{Efficiency comparisons. We report the and inference FPS of MROVSeg running on a RTX 3090. Training time is mesured with 2 NVIDIA H100.}
\vspace{-10pt} 
\label{tab:efficiency}
\setlength{\tabcolsep}{7pt}
\scriptsize
\begin{tabular}{cccccccc}
    \toprule
         Method  & Params. & \#GFLOPS & Inference FPS & Training Time  \\ 
        \hline \hline
      ZSSeg~\cite{simbase} &  530.8 & 22302.1 & 0.3 & 15h59min \\ 
      OVSeg~\cite{ovseg}    &  532.6  & 19345.6 & 0.4 & 17h54min \\ 
      CAT-Seg~\cite{catseg} &  433.7 &   2121.1 & 2.0 & \textbf{7h41min} \\
      FC-CLIP~\cite{fcclip} &  221.3   & 680.0 &  2.3 & 2d5h \\
      EBSeg~\cite{ebseg}  & 210.9   & 867.5 &  4.7 & 18h33min \\
      MROVSeg  & \textbf{162.1}  & \textbf{640.4} &  \textbf{10.5} & 9h40min  \\
      \bottomrule
    \end{tabular}
    \vspace{-5pt}
\end{table}

%% file: tables/ab_componets.tex
\begin{table}[t]
    \centering
    \setlength{\tabcolsep}{2.8pt}
    \caption{\textbf{Ablation study of components.} We show the effects of integrating each modules into the baseline.}
    \vspace{-5pt}
    \scriptsize
    \begin{tabular}{l|cccccc}
    \toprule
       Method  & PC-59 & A-150 & A-847 & PC-459 & VOC-20 \\ 
        \hline \hline
        Baseline & 52.3 &  25.4 &  9.0  & 10.8 & 92.9         \\
         + High-res Features & 54.1 & 28.1 & 10.4 & 13.0 & 93.6\\
         + Multi-Res Adapter   & 55.6 & 28.7 & 10.5 & 14.9 & 94.7\\
         + Masked Attention & 56.4 & 30.9 & 11.8 & 17.9 & 95.5 \\
         + Multi-grained Masked Attention & 58.7 &  32.4  & 12.9 & 19.2  & 95.8  \\
      \bottomrule
    \end{tabular}
    \vspace{-5pt}
    \label{tab:components}
\end{table}

%% file: tables/incremantal_comp.tex
\begin{table}[t]
    \centering
    \setlength{\tabcolsep}{2.8pt}
    \caption{\textbf{Effect of Hierarchical Mask Decoding and Image-conditioned Text Feature.}}
    \scriptsize
    \begin{tabular}{l|cccccc}
    \toprule
       Method  & PC-59 & A-150 & A-847 & PC-459 & VOC-20 \\ 
         \hline \hline
         MROVSeg w/o Hier. Mask Dec. & 57.1 &  30.5  & 11.3 & 18.0 & 95.1  \\
        MROVSeg w/o Img-cond. Text Feat. &58.5   &  32.0 & 12.1 & 19.6 & 95.5\\
        MROVSeg &58.7   &  32.4 & 12.9 & 19.2 & 95.8\\
      
      \bottomrule
    \end{tabular}
    \vspace{-10pt}
    \label{tab:inremental}
\end{table}

%% file: tables/ab_mradapter.tex
\begin{table}[t]
\centering
\caption{\textbf{Ablation study on various designs in Multi-Res Adapter.} The default setting are marked \underline{underline}.}
\resizebox{\linewidth}{!}{
\begin{tabular}{c|c|cccccc}
    \toprule
 \multirow{4}{*}{(a)} &  Spatial Fusion & A-847 & PC-459 & A-150 & PC-59 & VOC-20 
     \\
    \cmidrule{2-7}
  &  Concat. & 12.0& 18.4& 31.3& 57.1& 94.8\\
  & \underline{Depth.Conv.} & \textbf{12.9}& \textbf{19.2}& \textbf{32.4}& \textbf{58.7}& \textbf{95.8}\\
    \midrule  
 \multirow{4}{*}{(b)} &  Multi-Res Fusion & A-847 & PC-459 & A-150 & PC-59 & VOC-20 
     \\
    \cmidrule{2-7}
  &  Add & 12.0 & 17.7 & 31.2 & 55.9 & 94.5\\
  &  Concat. & 12.4 & \textbf{19.8} & 31.8 & \textbf{58.7} & 95.6\\
  &  \underline{Scale.Attn.} & \textbf{12.9}& 19.2& \textbf{32.4}& \textbf{58.7}& \textbf{95.8}\\
    \midrule    
    \multirow{4}{*}{(c)} &  Block Num & A-847 & PC-459 & A-150 & PC-59 & VOC-20
     \\
    \cmidrule{2-7}
  &  3 & 9.8 & 12.5 & 27.7 & 53.9 & 94.1\\
  &  \underline{6} & \textbf{12.9}& 19.2& \textbf{32.4}& 58.7& \textbf{95.8}\\
  &  9 & 12.7& \textbf{19.3}& 31.9& \textbf{59.2}& 95.4\\
    \midrule    
  \multirow{5}{*}{(d)} & Fusion Layer & A-847 & PC-459 & A-150 & PC-59 & VOC-20 
    \\
    \cmidrule{2-7}
   & \{3,6,9\} & 11.8& 19.1& 31.6& 56.5& \textbf{95.9}\\
   & \{3,6,9,12\} & 12.4 & \textbf{19.8}& 32.0 & 58.0& 95.4\\
   & \{stem,3,6,9\}& 12.8& 19.4& \textbf{32.6}& 58.1& 94.9\\
   & \underline{\{stem,3,6,9,12\}} & \textbf{12.9}& 19.2& 32.4& \textbf{58.7}& 95.8 \\
    \midrule
  \multirow{4}{*}{(e)} & Channel Width & A-847 & PC-459 & A-150 & PC-59 & VOC-20
     \\
    \cmidrule{2-7}
   & 384 & 10.4 & 17.0& 27.5& 56.1& 94.2\\
  &  \underline{768} & \textbf{12.9}& 19.2& 32.4& \textbf{58.7}& 95.8\\
  &  1024 & 11.1& 19.7& \textbf{31.0}& \textbf{59.5}& \textbf{96.1}\\
    \midrule
  \multirow{4}{*}{(f)} & Query Num & A-847 & PC-459 & A-150 & PC-59 & VOC-20
     \\
    \cmidrule{2-7}
   & \underline{100} & \textbf{12.9}& \textbf{19.2}& \textbf{32.4}& \textbf{58.7}& \textbf{95.8}\\
  &  200 & 11.9& 18.3& 31.8& 57.0& 93.8\\
  &  300 & 12.2& 18.7& \textbf{32.4}& 57.5& 95.2\\
    \bottomrule
\end{tabular}}
\label{tab:ab_mradapter}
\vspace{-15pt}
\end{table}

%% file: tables/convolution_conterpart.tex
\begin{table}[t]
    \centering
    \setlength{\tabcolsep}{3.9pt}
    \caption{\textbf{Comparison between different VLMs.} All Multi-grained Masked Attention is \textit{\textbf{disabled}} to adapt ConvNext.}
    \scriptsize
    \begin{tabular}{lc|ccccc}
    \toprule
       Method  & VLM & PC-59 & A-150 & A-847 & PC-459 & VOC-20 \\ 
         \hline \hline
          MROVSeg & ConvNext-L & 57.0 &  28.4  & 10.1 & 15.8 & 93.3  \\
        MROVSeg & ViT-L   &  58.5 & 30.4  &13.4& 20.1 & 95.4\\
      \bottomrule
    \end{tabular}
    \vspace{-10pt}
    \label{tab:conv_counterparts}
\end{table}

%% file: sec/sec_5_conclusion.tex
\vspace{-10pt}
\section{Conclusion}
In this paper, we introduce MROVSeg, a multi-resolution training framework designed to enhance open-vocabulary image segmentation by leveraging multi-resolution VLM features. The exceptional quantitative and qualitative results obtained on well-established open-vocabulary segmentation benchmarks serve as compelling evidence of its effectiveness and versatility. We hope our method can serve as a strong baseline for future research.